\documentclass[11pt,a4paper]{article}

\usepackage[comma,authoryear]{natbib}
\usepackage{amsmath}
\usepackage{amssymb}
\usepackage{hyperref}
\usepackage{url}
\usepackage{subcaption}
\usepackage{graphicx}

\newlength\titlebox
\title{Transfer learning from language models to image caption generators: Better models may not transfer better}

\author{
	Marc Tanti \qquad Albert Gatt\\
	Institute of Linguistics and Language Technology\\
	University of Malta\\
	marc.tanti.06@um.edu.mt \qquad albert.gatt@um.edu.mt
	\and
	Kenneth P. Camilleri\\
	Department of Systems and Control Engineering\\
	University of Malta\\
	kenneth.camilleri@um.edu.mt
}

\date{This is an unpublished pre-print. Comments are welcome.}

\begin{document}

\maketitle

\begin{abstract}
When designing a neural caption generator, a convolutional neural network can be used to extract image features. Is it possible to also use a neural language model to extract sentence prefix features? We answer this question by trying different ways to transfer the recurrent neural network and embedding layer from a neural language model to an image caption generator. We find that image caption generators with transferred parameters perform better than those trained from scratch, even when simply pre-training them on the text of the same captions dataset it will later be trained on. We also find that the best language models (in terms of perplexity) do not result in the best caption generators after transfer learning.
\end{abstract}

\section{Introduction}

Image caption generators make use of a convolutional neural network (CNN), typically pre-trained on a separate image-only dataset, in order to extract visual features from images. Using pre-trained neural networks to transform inputs into high level features for other neural networks makes it easier to avoid overfitting \citep{Vinyals2015}. Does this advantage only apply to the visual part of the caption generator? In this paper we investigate whether the language part of the image caption generator can also be handled by a pre-trained neural network that has been trained on a separate text-only dataset.

The language encoding part of a caption generator is the recurrent neural network (RNN) together with the embedding layer. We collectively call the parameters of these two layers `prefix encoding parameters' because they encode a partially generated caption prefix into a single vector in order to allow the softmax layer to predict the next word in the prefix. The source model from which we want to transfer these parameters is a trained neural language model and the target model is an untrained image caption generator.

\begin{figure}
	\centering
	\includegraphics[scale=0.8]{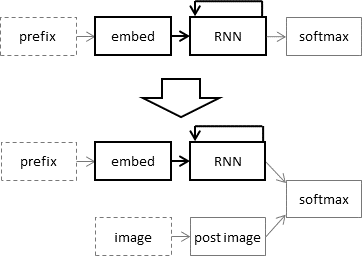}
	\caption{
		\label{fig:lt_architecture}
		The architectures of the language model (top) and caption generator (bottom) with the part that is transferred having a bold outline. Both neural networks take in a prefix of a sentence, embed its words, and encode it into a single vector via an RNN. In the case of the language model, this encoded prefix is then passed to a softmax layer which predicts the next word in the sentence prefix. In the case of the caption generator, the encoded prefix is first concatenated with an image feature vector before being passed to the softmax layer. The image feature vector starts as a 4\,096-element feature vector representing the image which is projected into a smaller vector via a fully connected layer (post image). This `post image' is what is concatenated to the prefix vector.
	}
\end{figure}

Not all caption generator architectures allow for this kind of parameter transferring. If the image is provided as an initial state to the RNN \citep{Devlin2015,Liu2016}, called an `init-inject' architecture \citep{Tanti2018}, then the image would need to be taken into account when training the RNN and hence cannot be trained separately. Instead we use a caption generator architecture called a `merge' architecture \citep{Tanti2018}, shown in Figure~\ref{fig:lt_architecture}, which leaves the vision encoding part and language encoding part of the caption generator separate \citep{Mao2014,Mao2015,Mao2015a}.

One challenge with transfer learning of language models is the vocabulary (for the softmax and embedding layer) which would differ between the language model's training set and the caption generator's training set. We set the vocabulary of the caption generator to be the intersection between the vocabulary extracted from the language model corpus and the vocabulary that would have been extracted from the captions dataset. This means that the final vocabulary of the caption generator would be smaller than that of a non-transferred caption generator. Note that this is what would happen in practice when an off-the-shelf language model is used to initialise a caption generator.

There are several factors that could affect the performance of the final caption generator, and rather than arbitrarily pick one configuration over others, we set our experiments to vary across three dimensions:
\begin{itemize}
	\item Domain: The corpus training data for the language model varies from data sampled from the image caption domain, to data sampled from general (news) text.

	\item Corpus size: The size of the corpus that is used to train the language model is varied from being one tenth to ten times the size of the captions dataset.

	\item Frozen prefix encoding parameters: After transferring the embedding layer and RNN parameters to the caption generator, they can either be frozen or fine tuned during training of the caption generator.
\end{itemize}

We compare the performance of the resulting caption generators with that of a non-transferred caption generator. We also compare the performance of the caption generators with that of the language models they are derived from. Following these experiments, a number of interesting observations were made:
\begin{itemize}
	\item Not only does transfer learning improve the caption generator's performance, even transferring from a language model that was pre-trained on the text of the exact same captions dataset (that is used to train the caption generator) results in better captions than initialising the parameters randomly.

	\item The performance of the caption generator does not improve monotonically with the size of the corpus used to train the language model; rather, peak performance tends to be reached somewhere before the largest size, with a subsequent decline when the training data is increased further.

	\item The source language model's performance is not perfectly correlated with the derived caption generator performance. It seems that there is a point where a language model can become too good to be used for transfer learning, perhaps because the internal representations used for language modelling are too specialised for use in other tasks.

	\item Notwithstanding the previous point, attempting to artificially vary the language model's performance by training for less epochs does not predictably affect the performance of the derived caption generator.

	\item The prefix encoding parameters of the caption generator can be set randomly and left frozen during training and the softmax layer will still learn to use these randomly specified (but deterministic) features, resulting in a decently performing caption generator.
\end{itemize}

Code used for these experiments is provided online.\footnote{See: \url{https://github.com/mtanti/mtanti-phd}.}

\section{Background}

Transfer learning \citep{Pan2010} is the act of exploiting the knowledge gained by a trained model in order to improve another model's learning process in a different task or domain. In NLP tasks, it is common to transfer word embeddings from other tasks, such as word2vec \citep{Mikolov2013} and GloVe \citep{Pennington2014}, in order to have more meaningful word vectors which were optimised using large text corpora. Apart from the word embedding layer, there is also work on transferring the RNN layer as well \citep{Zoph2016,Ramachandran2017,Howard2018,Mou2016}. \citet{Howard2018} claim that language modelling is a universal source task for transfer learning in natural language processing. They perform transfer learning by using a language model as an initial sentence encoder for classification tasks such as sentiment analysis, question classification, and topic classification. \citet{Ramachandran2017} also perform transfer learning from the source task of language modelling but transfer to machine translation and abstractive summarisation.

As in the present work, \citet{Ramachandran2017} implement an architecture that makes it possible to re-use the unconditioned source language model in a target task that requires a conditioned language model, which we solved by using the merge architecture. In their case, they made it possible to use an init-inject architecture by using two RNNs in series: The first RNN encoded the input text only. The second RNN was initialised with sentence vectors (for a translation task) but took as input the hidden state vectors of the first. Hence, the first RNN could be pre-trained. In our case, the merge architecture obviates the need for a dual RNN architecture, which is less computationally demanding.

Another important consideration in transfer learning is the relative value of freezing parameters versus fine-tuning during training. \citet{Yosinski2014} performed experiments on transfer learning in CNNs and tried transferring a variable number of layers from the input side of the neural network. With regards to the difference between freezing and fine-tuning, the first half of the layers tend to encode features that when frozen are difficult to exploit by randomly initialised later layers. Fine-tuning allows the transferred and randomly initialised layers to co-operate at reaching a suitable representation. This observation has been called fragile co-adaptation. \citet{Mou2016} found similar results for text classification tasks. When transferring the embedding layer and RNN, performance is always better when the transferred parameters are fine-tuned rather than left frozen. \citet{Mou2016} also found that when transferring between similar tasks, such as from a sentiment analysis task to a different sentiment analysis task, performance always improves when both the embedding layer and the RNN are transferred. On the other hand, when transferring between unrelated tasks, such as from a sentiment analysis task to a question classification task, only transferring the embedding layer without the RNN improves performance.

A third consideration in transfer learning is the relationship between the performance of the source model on the source task and the performance of the transferred model on the target task. In particular, we show below that the best performing language model does not transfer as well as lesser-performing models. \citet{Kornblith2018} found similar results with CNNs. Image features produced by different CNNs were used as inputs to logistic regressors that were trained to perform an image classification task. It turns out that the state-of-the-art CNNs do not produce the best fixed image features (according to the logistic regressor) but that it was some of the lesser CNNs that do. On a similar note, \citet{Hessel2015} found something similar with image caption generators. Different pre-trained CNNs were used to extract image features to be used for training image caption generators. It was found that using AlexNet \citep{Krizhevsky2012} results in slightly better captions than using VGG-16 \citep{Simonyan2014}, even though the first CNN has an object recognition top-1 accuracy of 57.1\% whilst the second's accuracy is 75.6\%. These results are evidence that as a neural network becomes better at performing the source task, its internal feature representations become overspecialised for the task and become less useful for performing other tasks.

Finally, the work of \citet{Hendricks2016} is similar to what we have done here. An image caption generator that uses the merge architecture (like we do) was developed in such a way that the vision handling part and the language handling part can be trained separately and then combined together. However, this work focuses on ways to extend the model's vocabulary after training time whereas we focus on treating the language handling part of a caption generator in the same way as the vision handling part: using a pre-trained neural network to extract fixed features.

\section{Experiments}

In these experiments, we train and evaluate image caption generators on Flickr8K \citep{Hodosh2013}. The rationale for doing so is that this is a relatively small dataset, which enables us to experiment with language models that are trained on corpora up to 10 times the number of sentences of the target corpus using our available hardware resources.

We train the neural language model on one of three different corpora. Each corpus' domain is of a varying degree of similarity to the final caption dataset's domain (Flickr8K). This allows us to see how the performance of the final caption generator changes as its parameters are transferred from more and more distant source domains. The three corpora are
\begin{itemize}
	\item Same captions: An in-domain corpus consisting of the sentences in the Flickr8K dataset itself, which is a performance ceiling since the source corpus cannot have a closer domain to the target corpus than the target corpus itself.
	
	\item Different captions: Another in-domain corpus consisting of the sentences in the MSCOCO \citep{Lin2014} dataset, which is another image captions corpus but which is different from Flickr8K.
	
	\item General text: An out-of-domain corpus consisting of the sentences in the Google one billion token language modelling benchmark corpus\footnote{See: \url{https://github.com/ciprian-chelba/1-billion-word-language-modeling-benchmark}} (LM1B), which is a corpus consisting mostly of news texts.
\end{itemize}

Flickr8K and MSCOCO were both obtained from the distributed versions provided by \citet{Karpathy2015}\footnote{See: \url{http://cs.stanford.edu/people/karpathy/deepimagesent/}}.

We also vary the size of the language model corpus training sets in order to measure the effect of size apart from domain, where sizes are measured as the number of sentences. A random sample of sentences from one of the above corpora is selected as a subcorpus. Each size of the corpus is computed as a multiple of the number of captions in Flickr8K (each caption is one sentence long), where the multiple is an exponent of 10. This allows us to measure how the performance of the caption generator changes as the corpus size is changed on a logarithmic scale, which gives us corpora sizes of 3\,000 sentences ($10^{-1}$th of Flickr8K), 9\,487 sentences ($10^{-0.5}$th of Flickr8K), 30\,000 sentences ($10^{0}$th of Flickr8K), 94\,868 sentences ($10^{0.5}$th of Flickr8K), and 300\,000 sentences ($10^{1}$th of Flickr8K).

Finally, we alternate between using frozen and fine-tuned transferred parameters. Freezing the parameters means that the prefix encoding parameters of the caption generator are not changed during training and are left as they were in the source language model. Fine-tuning the parameters means allowing them to be further optimised whilst training the caption generator.

Table~\ref{tbl:lt_configurations} shows all the different experimental configurations being compared. As a reference for comparing performance, we also train a caption generator on Flickr8K without using transfer learning (`no transfer').

\begin{table}
	\centering
	\begin{tabular}{c|cc|cc}
		 &	\multicolumn{2}{c|}{Language model} &	\multicolumn{2}{c}{Caption generator} \\
		Type &	Dataset &	Size multiple ($10^x$)  &	Dataset &	Frozen/Fine tuned \\
		\hline
		No transfer &	N/A &	N/A &	Flickr8K &	`Fine tuned' only \\
		Same captions &	Flickr8K &	-1, -0.5, 0 &	Flickr8K &	Frozen and fine-tuned \\
		Different captions &	MSCOCO &	-1, -0.5, 0, 0.5, 1 &	Flickr8K &	Frozen and fine-tuned \\
		General text &	LM1B &	-1, -0.5, 0, 0.5, 1 &	Flickr8K &	Frozen and fine-tuned \\
	\end{tabular}
	\caption{
		\label{tbl:lt_configurations}Experimental configurations that were compared. Size multiple refers to the number of sentences in the language model corpus such that a size multiple of $x$ means that the corpus has $10^x$th the number of sentences in Flickr8K ($10^x \times 30\,000$). Frozen/Fine tuned refers to whether the transferred parameters were frozen during training of the caption generator or allowed to be optimised together with the rest of the parameters.
	}
\end{table}

In all these corpora and in all the sizes, only words that occur at least 5 times in their respective training subcorpus were included in the vocabulary, with the remaining words being replaced by the unknown token. Since the caption generator can only work with words that were embedded by the language model, words in Flickr8K that were not in the language model corpus are also replaced by the unknown token (only the intersection of the two vocabularies was used in the caption generator whilst the full vocabulary was used in the language model).

All sentences were preprocessed by lowercasing all characters, replacing strings of digits with a single pseudo word, and removing all non-alphanumeric characters (excluding spaces). In order to reduce the memory requirements of running these experiments, the LM1B corpus was filtered so that only sentences that have no more than 50 tokens were included.

Early stopping was used on both language model training and caption generator training such that training is stopped on the epoch when the geometric mean of the perplexity on the validation set is worse than it was in the previous epoch. These validation sets were the ones provided with each dataset (Flickr8K, MSCOCO, and LM1B) which means that they are in the same domain as the training set. This is so that the language model would be closer to what would be expected from an off-the-shelf language model. Furthermore, the validation sets were not varied in size with the training sets. It is worth noting that in the case of `same captions', the validation set used for training the language model is the same as the one used for training the caption generator.

A gated recurrent unit \citep{Chung2014}, or GRU, was used as an RNN for both language models and caption generators. Biases were initialized to zero. Image features were extracted from the penultimate layer in the VGG-16 CNN \citep{Simonyan2014}.

\subsection{Hyperparameters}

The hyperparameters of each of the four different experimental configurations (rows shown in Table~\ref{tbl:lt_configurations}), both of the language model and of the derived caption generator, were tuned independently and automatically using Bayesian optimisation. In order to avoid spending too much time on tuning, only the training set size multiple of $10^0$ was used together with frozen prefix encoding parameters when tuning for each configuration. The rest of the variations on the same row in Table~\ref{tbl:lt_configurations} shared the same hyperparameters found. The prefix encoding parameters of the (trained) best language model (found whilst tuning its hyperparameters) are transferred to the caption generator when it is being tuned. The reason for choosing this method is that, in practice, the language model is tuned independently of the caption generation task, while the caption generator itself would be tuned to take advantage of the pre-trained language model. The `no transfer' model, on the other hand, has the caption generator being fully tuned without any restriction from a language model.

The library Scikit-Optimize\footnote{See: \url{https://scikit-optimize.github.io/}} was used to perform hyperparameter tuning using Bayesian optimisation. As an optimisation cost function, geometric mean of perplexity was used for the language model whilst the word mover's distance (WMD) metric \citep{Kusner2015,Kilickaya2017} was used on the caption generator.\footnote{The WMD metric was applied on captions generated from the resulting neural network when trained using the hyperparameters being evaluated.} The model, whose purpose is to predict the fitness of a given hyperparameter combination, is a random forest and was initialized using 32 random hyperparameter combinations paired with their evaluated fitness (the perplexity or WMD). The hyperparameter combinations were then optimised by exploring a sequence of 64 candidate hyperparameters that the model suggests will maximize the expected improvement in fitness. The best hyperparameters found for each configuration are shown in Table~\ref{tbl:lt_hyperparams}.

\begin{table}
	\centering
	\begin{subtable}{\textwidth}
		\centering
		\begin{tabular}{l|cccc}
			 &	No transfer &	Same captions &	Different captions &	General text \\
			\hline
			weight init. method &	N/A &	Xavier &	Normal &	Xavier \\
			max. init. weight &	N/A &	1.72e-01 &	8.25e-02 &	4.45e-01 \\
			embed size &	276 &	502 &	255 &	132 \\
			RNN size &	227 &	201 &	427 &	330 \\
			optimizer &	N/A &	RMSProp &	Adam &	Adam \\
			learning rate &	N/A &	2.49e-03 &	8.34e-04 &	5.98e-03 \\
			weight decay weight &	N/A &	3.72e-10 &	4.21e-08 &	1.45e-05 \\
			embedding dropout rate &	0.01 &	0.01 &	0.07 &	0.03 \\
			RNN dropout rate &	N/A &	0.33 &	0.13 &	0.23 \\
			max. gradient norm &	N/A &	6.96 &	7.54 &	47.90 \\
			minibatch size &	N/A &	210 &	104 &	68 \\
		\end{tabular}
		\caption{
			\label{tbl:lt_hyperparams_langmod}
			Hyperparameters for the language models.
		}
	\end{subtable}
	\vspace{10pt}

	\begin{subtable}{\textwidth}
		\centering
		\begin{tabular}{l|cccc}
			 &	No transfer &	Same captions &	Different captions &	General text \\
			\hline
			weight init. method &	Xavier &	Xavier &	Xavier &	Normal \\
			max. init. weight &	1.96e-01 &	2.43e-03 &	3.06e-04 &	4.52e-05 \\
			post image size &	268 &	430 &	511 &	307 \\
			post image activation &	ReLU &	ReLU &	ReLU &	none \\
			optimizer &	Adam &	RMSProp &	Adam &	Adam \\
			learning rate &	2.64e-04 &	2.83e-04 &	4.59e-05 &	1.30e-03 \\
			normalize image &	false &	false &	false &	true \\
			weight decay weight &	3.01e-07 &	2.45e-04 &	1.18e-10 &	2.87e-10 \\
			image dropout rate &	0.02 &	0.06 &	0.13 &	0.20 \\
			post image dropout rate &	0.21 &	0.29 &	0.01 &	0.31 \\
			RNN dropout rate &	0.28 &	0.41 &	0.18 &	0.01 \\
			max. gradient norm &	685.80 &	366.97 &	841.50 &	153.06 \\
			minibatch size &	237 &	227 &	18 &	162 \\
			beam width &	5 &	4 &	4 &	4 \\
		\end{tabular}
		\caption{
			\label{tbl:lt_hyperparams_capgen}
			Hyperparameters for the caption generators.
		}
	\end{subtable}
	\vspace{10pt}

	\caption{
		\label{tbl:lt_hyperparams}
		Best hyperparameters found for each experimental configuration. Initialization method is the probability distribution used to initialize all the weights (except transferred ones) and could be either the normal distribution or Xavier \citep{Glorot2010} with normal distribution. Max. init. weight is the maximum absolute value of the initial weight beyond which it is clipped and could be between 1e-5 and 1.0. Embed size, RNN size, and post image size all refer to the layer sizes in Figure~\ref{fig:lt_architecture} and could be between 64 and 512. Post image activation refers to the activation function used on the post image layer. The optimiser could be Adam \citep{P.Kingma2014}, RMSProp (\url{http://www.cs.toronto.edu/~tijmen/csc321/slides/lecture_slides_lec6.pdf}), or AdaDelta \citep{Zeiler2012} whilst the learning rate could be between 1e-5 and 1.0. Normalise image refers to whether to use the vector norm of the image feature vector or leave the image vector as-is. The weight decay weight could be between 1e-10 and 0.1 whilst the max. gradient norm could be between 1.0 and 1000.0. Dropout was used on the image vector itself, on the post image layer, on the embedding layer, and on the RNN hidden state vector and the dropout rate could be between 0.0 and 0.5. The minibatch size could be between 10 and 300 whilst the beam width could be between 1 and 5.
	}
\end{table}

\section{Results}

Each experiment was run five times, each time using a different randomly chosen subset of the corpus sentences to train the language model (as well as having different initial random weights, minibatches, and other non-deterministic values). The mean of the results for the quality of generated captions using METEOR \citep{Banerjee2005}, CIDEr \citep{Vedantam2015}, SPICE \citep{Anderson2016}, and WMD \citep{Kusner2015,Kilickaya2017} is shown in Table~\ref{tbl:lt_results_exp}.

\begin{table}
	\centering
	\begin{small}
		\begin{tabular}{ccr|cccc}
			Type &	Frozen? &	Size &	METEOR &	CIDEr &	SPICE &	WMD \\
			\hline
			no trans. &	no &	30000 &	{\underline{0.190}} (0.001) &	{\underline{0.457}} (0.011) &	{\underline{0.128}} (0.002) &	{\underline{0.137}} (0.003) \\
			\hline
			\hline
			same caps. &	yes &	3000 &	{{0.189}} (0.002) &	{{0.467}} (0.022) &	{{0.126}} (0.003) &	{{0.139}} (0.004) \\
			same caps. &	yes &	9487 &	{{0.193}} (0.001) &	{{0.476}} (0.015) &	{{0.131}} (0.002) &	{{0.140}} (0.002) \\
			same caps. &	yes &	30000 &	{\underline{0.194}} (0.002) &	{{0.476}} (0.010) &	{\underline{0.132}} (0.001) &	{{0.140}} (0.002) \\
			\hline
			same caps. &	no &	3000 &	{{0.190}} (0.004) &	{{0.447}} (0.017) &	{{0.127}} (0.003) &	{{0.137}} (0.003) \\
			same caps. &	no &	9487 &	{{0.194}} (0.001) &	\textbf{\underline{0.485}} (0.009) &	{{0.130}} (0.002) &	\textbf{\underline{0.141}} (0.002) \\
			same caps. &	no &	30000 &	{{0.194}} (0.003) &	{{0.478}} (0.011) &	{{0.131}} (0.003) &	{{0.140}} (0.003) \\
			\hline
			\hline
			diff. caps. &	yes &	3000 &	{{0.187}} (0.001) &	{{0.431}} (0.007) &	{{0.124}} (0.001) &	{{0.136}} (0.002) \\
			diff. caps. &	yes &	9487 &	{{0.189}} (0.001) &	{{0.451}} (0.006) &	{{0.126}} (0.002) &	{{0.139}} (0.001) \\
			diff. caps. &	yes &	30000 &	{{0.191}} (0.001) &	{{0.475}} (0.007) &	{{0.130}} (0.001) &	{{0.140}} (0.002) \\
			diff. caps. &	yes &	94868 &	{{0.192}} (0.002) &	{{0.478}} (0.009) &	{{0.132}} (0.002) &	{\underline{0.141}} (0.003) \\
			diff. caps. &	yes &	300000 &	{{0.190}} (0.001) &	{{0.469}} (0.010) &	{{0.130}} (0.001) &	{{0.139}} (0.002) \\
			\hline
			diff. caps. &	no &	3000 &	{{0.190}} (0.002) &	{{0.438}} (0.005) &	{{0.126}} (0.002) &	{{0.137}} (0.002) \\
			diff. caps. &	no &	9487 &	{{0.191}} (0.002) &	{{0.459}} (0.013) &	{{0.129}} (0.002) &	{{0.138}} (0.002) \\
			diff. caps. &	no &	30000 &	\textbf{\underline{0.195}} (0.002) &	{{0.482}} (0.010) &	\textbf{\underline{0.133}} (0.002) &	{{0.140}} (0.003) \\
			diff. caps. &	no &	94868 &	{{0.194}} (0.002) &	{{0.471}} (0.013) &	{{0.132}} (0.002) &	{{0.137}} (0.002) \\
			diff. caps. &	no &	300000 &	{{0.194}} (0.002) &	{\underline{0.482}} (0.008) &	{{0.133}} (0.001) &	{{0.140}} (0.002) \\
			\hline
			\hline
			gen. text &	yes &	3000 &	{{0.145}} (0.006) &	{{0.245}} (0.017) &	{{0.071}} (0.005) &	{{0.095}} (0.004) \\
			gen. text &	yes &	9487 &	{{0.171}} (0.004) &	{{0.364}} (0.015) &	{{0.113}} (0.003) &	{{0.123}} (0.003) \\
			gen. text &	yes &	30000 &	{{0.182}} (0.002) &	{{0.425}} (0.004) &	{{0.122}} (0.001) &	{{0.134}} (0.001) \\
			gen. text &	yes &	94868 &	{{0.183}} (0.002) &	{{0.446}} (0.011) &	{{0.125}} (0.002) &	{{0.135}} (0.003) \\
			gen. text &	yes &	300000 &	{{0.186}} (0.002) &	{\underline{0.453}} (0.010) &	{{0.127}} (0.001) &	{\underline{0.137}} (0.002) \\
			\hline
			gen. text &	no &	3000 &	{{0.156}} (0.003) &	{{0.220}} (0.011) &	{{0.074}} (0.002) &	{{0.097}} (0.002) \\
			gen. text &	no &	9487 &	{{0.183}} (0.002) &	{{0.370}} (0.008) &	{{0.118}} (0.003) &	{{0.124}} (0.002) \\
			gen. text &	no &	30000 &	{{0.187}} (0.001) &	{{0.419}} (0.011) &	{{0.125}} (0.001) &	{{0.130}} (0.003) \\
			gen. text &	no &	94868 &	{{0.187}} (0.001) &	{{0.431}} (0.015) &	{{0.125}} (0.002) &	{{0.133}} (0.004) \\
			gen. text &	no &	300000 &	{\underline{0.190}} (0.002) &	{{0.440}} (0.012) &	{\underline{0.128}} (0.002) &	{{0.134}} (0.002) \\
			\hline
		\end{tabular}
	\end{small}
	\caption{
		\label{tbl:lt_results_exp}
		Results for the final generated captions after transfer learning. Underlined values are the best results for each experiment type whilst boldfaced values are the best results across all types. Legend: no trans. - no transfer learning, frozen - frozen parameters (vs. fine-tuned), size - corpus size, diff. caps. - different captions, gen. text - general text.
	}
\end{table}

\subsection{Transfer learning versus non-transfer learning}

Transfer learning always improves over non-transfer learning, although not drastically. Fine-tuning tends to improve the scores more than freezing, but it does so by a very small margin, meaning that the prefix encoding parameters are transferable between language modelling and caption generation as-is. Regarding the language model corpus, domain plays an important role: the general text corpus (LM1B) never performs better than an in-domain corpus with 9\,487 sentences, even when 300\,000 sentences are used. In fact, when 300\,000 sentences are used to train the general text language model, the resulting caption generator performance is on a par with the performance obtained without using any transfer learning.

It is interesting that simply pre-training the prefix encoding parameters on the text of the same captions dataset that will be used to train the caption generator will improve the performance of the final caption generator. This fact could be of great practical importance when training neural networks, possibly as a form of smart initialization where the caption generator's prefix encoding parameters are initialized at a sensible point in parameter space. It could be argued that this is instead the result of better hyperparameter tuning due to a transferred caption generator having less hyperparameters to optimise (the embedding and RNN sizes are determined and fixed by the source language model whilst the caption generator needs to optimise them as well), but this is in fact a practical advantage of transfer learning where the dimensionality of the hyperparameter search space is reduced.

\subsection{Size of language model corpus}

\begin{figure}
	\centering
	\begin{subfigure}{0.45\textwidth}
		\includegraphics[scale=0.7]{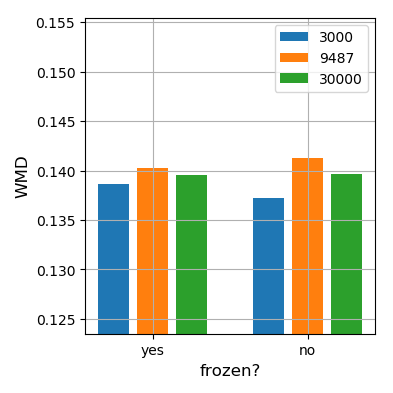}
		\caption{
			\label{results_wmd_samecaps}
			In-domain: same captions
		}
	\end{subfigure}
	\quad
	\begin{subfigure}{0.45\textwidth}
		\includegraphics[scale=0.7]{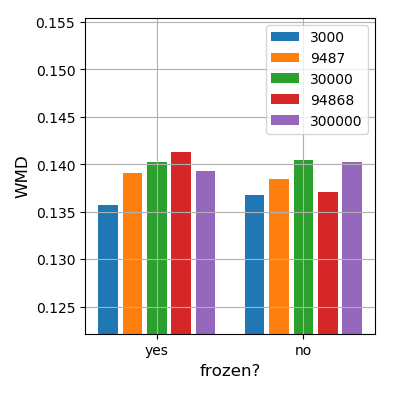}
		\caption{
			\label{results_wmd_diffcaps}
			In-domain: different captions
		}
	\end{subfigure}
	\vspace{10pt}
	
	\begin{subfigure}{0.45\textwidth}
		\includegraphics[scale=0.7]{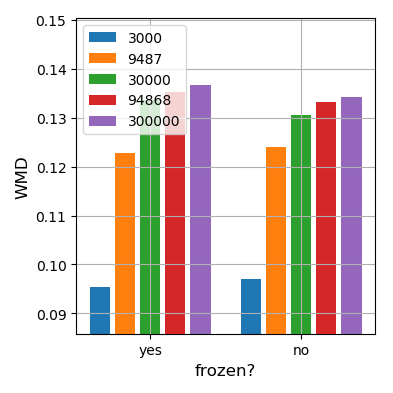}
		\caption{
			\label{results_wmd_gentext}
			Out-of-domain: general text
		}
	\end{subfigure}
	\caption{
		\label{fig:results_wmd}
		The WMD score of the different language models measured on the Flickr8K test set captions. Each colour bar shows the number of sentences that the language model was trained on. Note how the largest in-domain corpora do not result in the best WMD. While this is not the case in the out-of-domain corpus, it also does not perform as well as the in-domain corpora.
	}
\end{figure}

Why doesn't increasing the language model corpus size always increase the performance? In the case of the `same captions' and `different captions' models, on most quality metrics, pre-training on part of the language model corpus gives a better performance than pre-training on the whole. Figure~\ref{fig:results_wmd} shows more clearly how the caption generator's WMD score changes as the language model's corpus size changes.

Is the reason related to the language model's performance somehow? We measured the geometric mean of the perplexity of the language models on their respective validation set corpus in order to measure the performance of the language model rather than the caption generator.

One must be careful when doing this since the different language models have different vocabulary sizes and the unknown token gives an unfair advantage to smaller vocabularies. To understand this, imagine if, in the extreme case, all the words were omitted from the vocabulary and were all replaced with the unknown token. This would make every word almost perfectly predictable (save for the end of sentence token) and the language model would assign almost 100\% of the probability to the unknown token each time, leading to a perplexity that is almost perfect. Adding words to the vocabulary would then make prediction more uncertain and thus lead to a worse perplexity, with larger vocabularies containing more uncertainty than smaller ones. To get around this problem and make the unknown token behave more fairly, we divide the probability assigned to the unknown token by the number of different words it replaces in the corpus we are using it on. For example, if the number of different words in the corpus being used to evaluate the language model is 1\,000 and the known vocabulary covers 400 of those word types, then the unknown token will be replacing the remaining 600 word types. Given that the language model does not give any information about those 600 word types, we assume that they are uniformly distributed and assume that the probability assigned by the language model to the unknown token is evenly divided between all those 600 word types. Now, whenever we encounter an unknown token in the corpus, we replace its probability $p$ by $\frac{p}{600}$. This effectively makes the vocabulary size equal to 1000 again, with the 600 out-of-vocabulary words dividing the unknown token's probability among themselves. This eliminates the advantage gained by smaller vocabularies as all vocabularies are of equal size now. Note that this is only fair if all language models are being evaluated on the same corpus (which they are) as otherwise the vocabulary size will not be fixed among all language models.

\begin{figure}
	\centering
	\begin{subfigure}{0.45\textwidth}
		\includegraphics[scale=0.7]{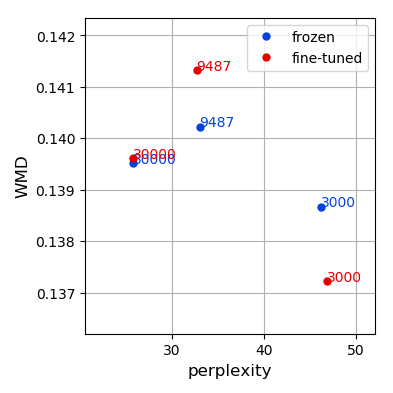}
		\caption{
			\label{results_pplx-wmd_samecaps}
			In-domain: same captions
		}
	\end{subfigure}
	\quad
	\begin{subfigure}{0.45\textwidth}
		\includegraphics[scale=0.7]{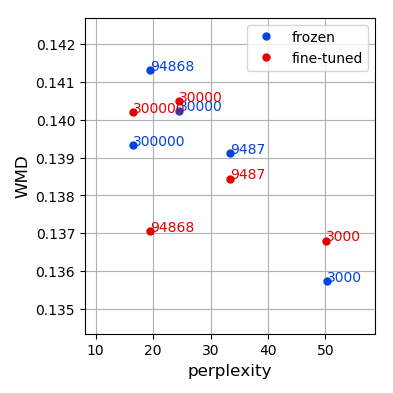}
		\caption{
			\label{results_pplx-wmd_diffcaps}
			In-domain: different captions
		}
	\end{subfigure}
	\vspace{10pt}
	
	\begin{subfigure}{0.45\textwidth}
		\includegraphics[scale=0.7]{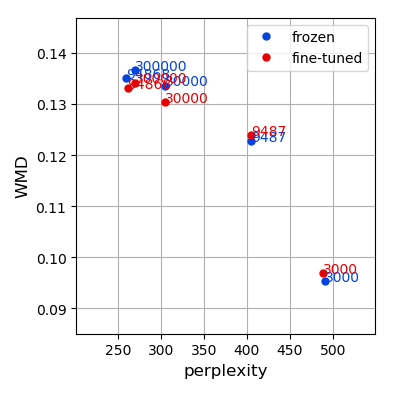}
		\caption{
			\label{results_pplx-wmd_gentext}
			Out-of-domain: general text
		}
	\end{subfigure}
	\caption{
		\label{fig:results_pplx-wmd}
		Scatter plot relating language model perplexity and transferred caption generator WMD score. To the right of each point is the number of sentences used to train the language model. Points that are vertically higher have a better WMD score whilst points that are further to the left have a better perplexity. Note how, in general, the best perplexity does not give the best WMD score.
	}
\end{figure}

After comparing the perplexities to the WMD scores, it turns out the language models generally perform better as the corpus size increases and, as a consequence, the best perplexity does not result in the best WMD score after transfer learning. A scatter plot showing the relationship between language model perplexity and transferred caption generator WMD is shown in Figure~\ref{fig:results_pplx-wmd}. The WMD score starts improving with every improvement in perplexity until the best perplexity is reached, at which point the score dips. There seems to be an exception in the case of the fine-tuned version of the `different captions' model where an outlier seems to disrupt the trend but everywhere else the trend is preserved. There also seems to be another anomaly in the `general text' model where the best perplexity does not come from the largest corpus but from the second largest one. This could be due to the vocabulary being way too large for the hyperparameters chosen for the language model (which were tuned using a smaller corpus).

So why does the WMD score stop correlating with the corpus size after a point? We explain this as a case of overspecialisation. A language model that is too good at language modelling will produce an internal representation of the sentence prefix that is too specialised to language modelling to be also useful (and hence transferable) to the caption generation task. We are thus in agreement with \citet{Kornblith2018} that state-of-the-art neural networks in a source task might not be ideal to perform transfer learning in a target task.

\subsection{Partial training of language model}

The previous results led us to a new hypothesis that, rather than varying the perplexity of the language model by varying the training corpus sizes, we can instead prematurely stop the language model's training process before peak validation performance is reached and check if this will also lead to better transferability. We tested this hypothesis by partially training the language model for a set number of epochs before transferring the prefix encoding parameters and measuring the resulting caption generator's WMD score.

Given a number of epochs $n$, we trained the language model for $n$ epochs and then transferred its prefix encoding parameters to the caption generator. We varied $n$ to be between 0 (no language model training, just transfer the random parameters as-is) and 15. For each $n$, we retrain both the language model and caption generator for five times, just like the main experiments described above.

We only accept a trained language model if its perplexity on the validation set kept improving after every epoch. If not, we started training over again. If after five attempts at re-training a language model, the validation perplexity keeps peaking before reaching the $n$th epoch, we terminate training there, and record at which $n$ this happened. We then continue increasing $n$ without early stopping in order to see if an overfitted language model (overfitted according to the validation set) results in a better or worse caption generator. The largest $n$ we reached with early stopping (no overfitting) was 13.

For each language model corpus, we included both frozen and fine-tuned prefix encoding parameters. Since this experiment takes a long time to complete, we only used one corpus size which is 30\,000 sentences, that is, the full size of Flickr8K, with the other corpora taking a random sample of sentences as described before. The results, shown in Figure~\ref{fig:results_partialtrain_wmd}, reveal several interesting points:

\begin{figure}
	\centering
	\begin{subfigure}{0.45\textwidth}
			\includegraphics[scale=0.7]{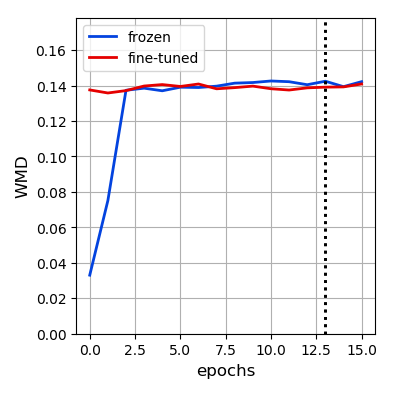}
			\caption{
				\label{results_partialtrain_wmd_samecaps}
				In-domain: same captions. Best WMD (0.142) reached at epoch 10 with frozen prefix encoding parameters. Overfitting occurred after epoch 13.
			}
		\end{subfigure}
		\quad
		\begin{subfigure}{0.45\textwidth}
			\includegraphics[scale=0.7]{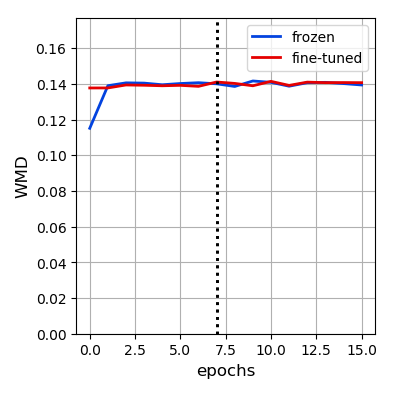}
			\caption{
				\label{results_partialtrain_wmd_diffcaps}
				In-domain: different captions. Best WMD (0.141) reached at epoch 9 with frozen prefix encoding parameters. Overfitting occurred after epoch 7.
			}
		\end{subfigure}
		\vspace{10pt}
		
		\begin{subfigure}{0.45\textwidth}
			\includegraphics[scale=0.7]{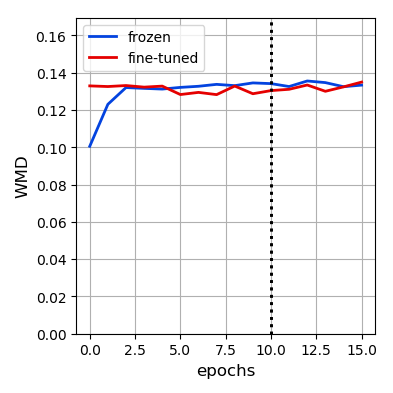}
			\caption{
				\label{results_partialtrain_wmd_gentext}
				Out-of-domain: general text. Best WMD (0.136) reached at epoch 12 with frozen prefix encoding parameters. Overfitting occurred after epoch 10.
			}
		\end{subfigure}
	\caption{
		\label{fig:results_partialtrain_wmd}
		How the WMD score in the transferred caption generator changes as the language model is trained for a varying number of epochs. The dark vertical line shows the last epoch before the language model started overfitting (in terms of perplexity measured on the validation set).
	}
\end{figure}

\begin{itemize}
	\item Both the frozen and fine-tuned versions started overfitting together (same epoch) for all language model corpora.
	
	\item After training for two epochs, the transferred frozen parameters are sufficient for generating captions that are close to the best quality.
	
	\item The best frozen parameters version is slightly superior to the best fine-tuned version for all three language model corpora. In fact, the best result here, which was obtained with the frozen version, is slightly better than the best result for the experiments in Table~\ref{tbl:lt_results_exp} which happens to have been obtained with the fine-tuned version (0.141 as opposed to 0.142 here).
	
	\item For the `different captions' and `general text' corpora, using the randomly initialised prefix encoding parameters and leaving them frozen (frozen zero epoch training) does not result in substantially degraded performance. It is important to understand that although an RNN has random weights, it does not act non-deterministically. In fact it still encodes something from the prefix of the partially generated caption; it's just that the encoding is not optimised to work well on the training set. The neural layer that reads and processes the RNN's hidden state vector can still find some useful features about the prefix from the `wild' encoding given by a random RNN. This recalls findings in research on echo state networks \citep{Jaeger2004}, where a randomly initialised simple RNN is frozen during training and only the layer that reads the RNN's hidden state vector is trained. With echo state networks, however, you are supposed to give the RNN's weights a spectral radius that is less than one, whereas here we don't use any such restriction.
	
	\item Although the `same captions' corpus does result in substantially degraded performance for the frozen zero epoch training version, we determined that this is due to differences in the chosen hyperparameters. In fact, we found that a single hyperparameter change can give substantial improvement to the frozen zero epoch training version: initializing weights using a normal distribution instead of using Xavier initialization. This results in a WMD of 0.118, exceeding the WMD of 0.115 of `different captions'. Xavier initialization uses a smaller variance in random values which probably results in most of the activations in the RNN's hidden state vector being similar and hence making it harder to accidentally encode enough useful distinct features. That said, the frozen zero epoch training version of the `general text' language model still works relatively well with Xavier initialization so this explanation is not complete.
	
	\item Finally, and most importantly, partial training does not seem to have the same effect as limiting the amount of data to train on. Whereas the previous results showed that changing the amount of data results in a predictable change in performance, changing the amount of epochs for which to train the language model seems to result in a more haphazard change in performance.
\end{itemize}

\section{Conclusion}

We have shown that a caption generator benefits from having its embedding layer and RNN transferred from a language model, but mostly when transferring from an in-domain corpus. It also benefits from simply pre-training the embedding layer and RNN on the text of the same captions dataset that it will eventually be trained on after transferring. We have also shown that this will only work if the language model is only trained on a fraction of the dataset. Beyond a certain number of sentences in the training corpus, the internal features produced by the language model might become overspecialised to the task of language modelling, which interferes with the caption generation task. Furthermore, partial training of the language model does not seem like a reliable way to prevent this overspecialisation from happening.

For future work we would like to try training a caption generator on MSCOCO after having pre-trained the prefix encoding parameters on the text of MSCOCO itself. It would also be interesting to see if there is a way to make CNNs produce more generic image features that are more useful for caption generators by limiting the amount of images they are trained on. Finally, does this `better models not transferring better' observation hold in general? A standardised series of experiments on different machine learning tasks would be beneficial.

\section*{Acknowledgments}

The research in this paper is partially funded by the Endeavour Scholarship Scheme (Malta). Scholarships are part-financed by the European Union - European Social Fund (ESF) - Operational Programme II - Cohesion Policy 2014-2020 “Investing in human capital to create more opportunities and promote the well-being of society”.

\bibliographystyle{apalike}
\bibliography{ms}

\end{document}